\RequirePackage{fix-cm}
\documentclass[smallextended]{svjour3}       
\smartqed  
\usepackage[english]{babel}
\usepackage{graphicx}
\usepackage{amsmath}
\usepackage{amssymb}
\usepackage{mathtools}
\usepackage{float}
\usepackage[linesnumbered, ruled, vlined]{algorithm2e}
\usepackage{multirow}
\usepackage{url}

\usepackage{array}
\newcolumntype{L}[1]{>{\raggedright\let\newline\\\arraybackslash\hspace{0pt}}m{#1}}
\newcolumntype{C}[1]{>{\centering\let\newline\\\arraybackslash\hspace{0pt}}m{#1}}
\newcolumntype{R}[1]{>{\raggedleft\let\newline\\\arraybackslash\hspace{0pt}}m{#1}}

\journalname{Soft Computing}

\begin{document}


\title{Solving Complex Multi-UAV Mission Planning Problems using Multi-objective Genetic Algorithms
}


\titlerunning{Solving Complex Multi-UAV MPPs using MOGAs}         

\author{Cristian Ramirez-Atencia \and Gema Bello-Orgaz \and Mar\'ia D. R-Moreno \and David Camacho \and}


\institute{Cristian Ramirez-Atencia \and Gema Bello-Orgaz \and David Camacho \at
			 Departamento de Inform\'atica, Universidad Auton\'onoma de Madrid\\
			 \email{cristian.ramirez@inv.uam.es}, \email{gema.bello@uam.es}, \email{david.camacho@uam.es}
             \and
             Mar\'ia D. R-Moreno \at
             Departamento de Autom\'atica, Universidad de Alcal\'a \\
             \email{mdolores@aut.uah.es}
}

\date{Received: date / Accepted: date}

\maketitle

\begin{abstract}
Due to recent booming of UAVs technologies, these are being used in many fields involving complex tasks. Some of them involve a high risk to the vehicle driver, such as fire monitoring and rescue tasks, which make UAVs excellent for avoiding human risks. Mission Planning for UAVs is the process of planning the locations and actions (loading/dropping a load, taking videos/pictures, acquiring information) for the vehicles, typically over a time period. These vehicles are controlled from Ground Control Stations (GCSs) where human operators use rudimentary systems. This paper presents a new Multi-Objective Genetic Algorithm for solving complex Mission Planning Problems (MPP) involving a team of UAVs and a set of GCSs. A hybrid fitness function has been designed using a Constraint Satisfaction Problem (CSP) to check if solutions are valid and Pareto-based measures to look for optimal solutions. The algorithm has been tested on several datasets optimizing different variables of the mission, such as the makespan, the fuel consumption, distance, etc. Experimental results show that the new algorithm is able to obtain good solutions, however as the problem becomes more complex, the optimal solutions also become harder to find.
\keywords{Unmanned Air Vehicles, Mission Planning, Multi-Objective Optimization, Genetic Algorithms, Constraint Satisfaction Problems}
\end{abstract}

\section{Introduction}\label{introduction}

Nowadays, Unmanned Air Vehicles (UAVs) or drones have become very popular in many potential applications including surveillance \cite{PereiraEtAl2009}, disaster and crisis management \cite{WuEtAl2006}, and agriculture or forestry \cite{MerinoEtAl2006} among others. For this reason, many research works related to this field have been developed over the past 20 years \cite{Kendoul2012} \cite{LeeEtAl2008} \cite{Rodriguez-FernandezEtAl2015d}.

The rapid development of the vehicles capabilities has caused their incorporation into many areas to perform complex tasks which involve a high risk to the vehicle driver, such as detecting forest fires or rescue tasks. So using UAVs avoid risking human lives while their manageability permits to reach areas of hard access.

The process of mission planning for a team of UAVs involves generating tactical goals, commanding structure, coordination, and timing. Currently, UAVs are controlled remotely by human operators from Ground Control Stations (GCSs), using rudimentary planning systems, such as following preplanned or manually provided plans. In order to perform more complex tasks and coordinated missions, these systems require more advanced capabilities.

Mission planning problems (MPPs) are a big challenge in actual NP-hard optimization problems. Classic planners are based on graph search or use a logic engine. But this kind of planners have several limitations, probably the most important is the big computational cost that their algorithms need to solve these missions. These missions have a lot of requirements that need to be considered and it is also necessary to coordinate all the UAVs. These requirements generate search graphs that need a huge process capabilities to find a solution. In addition, Multi-UAV missions usually require the use of several GCSs for controlling all the UAVs involved. This generates a new Multi-GCS approach that makes this problem even harder to solve.

Another critical issue in MPP is that there are several parameters which can be used to define the quality of a solution, such as the fuel consumption, the makespan, the cost of the mission, etc. In these cases, a Pareto Optimal Frontier (POF) can be computed in order to get the best solutions optimizing different objectives at the same time. Due to mission planning is based on search problems, an option to solve this type of problems could be using Multi-Objective Evolutionary Algorithms (MOEAs). In this work, we extend a previous work \cite{Bello-OrgazEtAl2015} in order design and implement a Multi-Objective Genetic Algorithm (MOGA) to solve this problem. For this purpose, a fitness function consisting of two phases has been designed. Firstly, modelling the MPP as a CSP, the fitness function checks that the solution plans fulfill all the constraints given by the different capabilities of the UAVs and the GCSs involved. Afterwards, using the validated  plans, a Pareto-based function is calculated to optimize different quality parameters of the solutions.

The rest of the paper is structured as follows. Section \ref{relatedwork} describes the related work concerning mission planning, CSPs and GAs. Section \ref{uavsmissionplans} presents the Misison planning problem, while section \ref{cspmodelling} presents the CSP approach used to model it. Section \ref{mogaapproach} presents the MOGA-CSP approach, the encoding designed and the fitness function implemented to solve Multi-GCS MPP. Section \ref{experiments} provides a description of the dataset employed, the setup employed in the MOGA-CSP and a complete experimental evaluation of it. Finally, in Section \ref{conclusions} the conclusions and some future research lines of the work are presented.

\section{Related Work}\label{relatedwork}

This section starts with a general introduction to Mission Planning techniques. After this brief introduction, an overview of Constraint Satisfaction Problems is presented showing the different methods used in the literature to solve them. Finally, a description of Genetic Algorithms (GAs) and their applications to optimization problems has been carried out.

\subsection{Mission Planning}\label{missionplanning}

Planning has been an area of research in Artificial Intelligence (AI) for over three decades. A variety of tasks including robotics \cite{DiazEtAl2013}, web-based information gathering \cite{KuterEtAl2005}, autonomous agents \cite{CamachoEtAl2006_2} and mission control \cite{VachtsevanosEtAl2005} have benefited from planning techniques. Moreover, mission planning is a common problem in AI. A mission can be described as a set of goals that are achieved by performing some task with a group of resources over a period of time. The whole problem can be summed up in finding the correct schedule of resource-task assignments that satisfies the proposed constraints.

In the literature there are some attempts to implement mission planning systems. Doherty et al. \cite{DohertyEtAl2009} presents an architectural framework for Mission Planning and execution monitoring, using temporal action logic (TAL). Fabiani et al. \cite{FabianiEtAl2007} modelled the problem for search and rescue scenarios using Markov Decision Process (MDP) and solve it with dynamic programming algorithms. German Aerospace Centre (DLR) also developed a mission management system based on the behaviour paradigm \cite{AdolfEtAl2010} which has been integrated onboard the ARTIS helicopter and validated in different scenarios, including waypoints following and search and track missions.

An essential concept in Mission Planning is cooperation or collaboration, which occurs at a higher level when various UAVs work together in a common mission sharing data and controlling actions together. There are few contributions that deal with Multi-UAV problems in a deliberative paradigm (cooperative task assignment and mission planning). Bethke et al. \cite{BethkeEtAl2008} proposed an algorithm for cooperative task assignment that extends the receding-horizon task assignment (RHTA) algorithm to select the optimal sequence of tasks for each UAS. Another approach by Kvarnstrom et al. \cite{KvarnstromEtAl2010} propose a new mission planning algorithm for collaborative UAVs based on combining ideas from forward-chaining planning with partial-order planning. This approach led to a new hybrid partial-order forward-chaining (POFC) framework that meets the requirements on centralization, abstraction, and distribution found in realistic emergency services settings.

Finally, other approaches formulate the mission planning problem as a Constraint Satisfaction Problem (CSP), where the tactic mission is modelled and solved using constraint satisfaction techniques \cite{Ramirez-AtenciaEtAl2015b}.

\subsection{Constraint Satisfaction Problems}\label{csps}

The MPP can be summed up in finding the correct schedule of resource-task assignments which satisfies the proposed constraints, like a CSP. It can be defined as follows \cite{Bartak1999}:

\begin{itemize}
\item A set of variables $V={v_1,...,v_n}$.
\item For each variable, a finite set of possible values $D_i$ (its domain).
\item A set of constraints $C_i$ restricting the values that variables can simultaneously take.
\end{itemize}

A CSP is usually represented as a graph where the pairs $<$Variable,Value$>$ are the nodes and the constraints are the edges. Although there are other representations as those presented in \cite{Gonzalez-PardoEtAl2014_1} for Ant Colony Optimization and videogames. In the literature, there are many proposed methods to search the space of solutions for CSPs, such as Backtracking (BT), Backjumping (BJ) or look-ahead techniques (i.e. Forward Checking (FC) \cite{BessiereEtAl1999}) among others. These algorithms are usually combined with other techniques like consistency techniques \cite{Bessiere2006} (domain consistency, arc consistency or path consistency) to modify the CSP and ensure its local consistency conditions.

In many real-life applications it is necessary to find a good solution, and not the complete space of possible solutions. For this purpose, combining a CSP with an optimization function results in a Constraint Satisfaction Optimization Problem (CSOP). In these approaches the optimization function maps every solution (complete labelling of variables) to a numerical value measuring the quality of the solution. The most widely used algorithm for finding optimal solutions is called Branch \& Bound (B\&B) \cite{RasmussenEtAl2006}. This algorithm searches for solutions in a depth first manner pruning the sub-tree under the current partial labelling when it exceeds the bound of the best value so far. In the case of Multi-Objective Optimization, an extension of this method, known as Multi-objective Branch \& Bound (MOBB) \cite{Rodriguez-FernandezEtAl2015a} is used to find the Pareto Optimal Frontier (POF) composed of all non-dominated solutions of the problem. Other methods for solving CSOP include Russian doll search \cite{RollonEtAl2007}, Bucket elimination \cite{RollonEtAl2006}, Genetic algorithms \cite{FonsecaEtAl1998} and Swarm intelligence \cite{Gonzalez-PardoEtAl2013}.

A TCSP is a particular class of CSP where variables represent times (time points, time intervals or durations) and constraints represent sets of allowed temporal relations between them \cite{SchwalbEtAl1998}. Different classes of constraints are characterized by the underlying set of basic temporal relations (BTR). Most types of TCSPs can be represented using Point Algebra (PA), with  BTR = $\{\varnothing, <, =, >, \le, \ge, ?\}$. A commonly used approach is Allen's Interval Algebra \cite{Allen1983}, which defines several relations between time intervals, with BTR = $\{<, >, m, mi, o, oi, s, si, d, di, f, fi, =\}$.

In the related literature, Mouhoub \cite{Mouhoub2002} proved that on real-time or Maximal TCSPs (MTCSPs), the best methods for solving TCSPs are Min-Conflict-Random-Walk (MCRW) for under-constrained and middle-constrained problems, and Tabu search and Steepest-Descent-Random-Walk (SDRW) in the over-constrained case. In this work, the author also developed a temporal model (TemPro \cite{Mouhoub2004}) which was based on interval algebra, to translate an application involving temporal information into a CSP. A TCSP can perfectly represent a Multi-UAV mission as a set of temporal constraints over the time the tasks in the mission start and end. 

\subsection{Genetic Algorithms}\label{gas}

Genetic Algorithms (GAs) have been traditionally used in a large number of different domains, mainly related to optimization problems \cite{Holland1992}. These stochastic methods are inspired by natural evolution and genetics, and the complexity of the algorithm depends on the codification and the operations used to reproduce, cross, mutate and select the different individuals of the population.

There is a wide range of applications where GAs have been successful, from optimization \cite{BinEtAl2010} to Data Mining \cite{Bello-OrgazEtAl2014} \cite{MenendezEtAl2014}. GAs have demonstrated to be robust, able to find satisfactory solutions in highly multidimensional problems with complex relationships between the variables. In recent works \cite{HaoEtAl2015} \cite{Ramirez-AtenciaEtAl2015a}, GAs have been used to represent CSPs.

Regarding the application of GAs to solve MPPs, there are several works in the literature. The Soliday et al. \cite{SolidayEtAl1999} approach developed a GA to solve UAV missions under complex constraints. The GA was constructed using a novel representation based on the nearest neighbour search, being each allele the N nearest neighbour, and uses a qualitative fitness function based on the number of mission objectives and the time permitted. Tang \cite{TangEtAl2011} created a nested GA for military planning (resource allocation and task scheduling) based on the robustness measure (RM) and test it with different probabilities and durations. In Geng et. al. \cite{GengEtAl2013} work, the authors designed a graph based representation for mission planning of UAVs to carry out a series of tasks. The flying space for these tasks was constrained with the presence of flight prohibited zones (EPZs) and enemy radar sites. Finally, in Savurant et. al. \cite{SavuranEtAl2015}, authors presented a GA for the Capacity Mobile Depot Vehicle Routing Problem, improving the GA process using Insertion Local Search (ILS) and 2-opt local search.

In MPPs for Multi-UAVs can be taken into account several criteria to measure the quality of a solution, such as the fuel consumption, the makespan or the cost of the mission, among others. Therefore, it can be interesting to optimize simultaneously different objectives in order to get the best solutions. This type of problems could solved using Multi-Objective Genetic Algorithms (MOGAs) \cite{ZhouEtAl2011} \cite{ZitzlerEtAl2004} based on Pareto optimization techniques. The most known approaches are SPEA2 \cite{DebEtAl2002} and NSGA-II \cite{ZitzlerEtAl2001}.

\begin{figure}[!h]
		\includegraphics[width=0.75\textwidth]{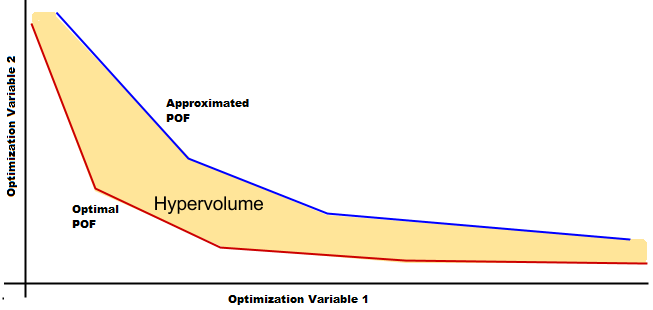}
		\centering
		\caption{Hypervolume for two optimization variables. The optimal POF is represented in red, the solutions obtained using a specific algorithm are represented in blue, and the hypervolume comprised between them is represented in yellow.}
		\label{fig:hypervolumeExample}
	\end{figure}

Finally, in order to evaluate the performance of the algorithms, there are some performance metrics such as the Hypervolume \cite{ZitzlerEtAl2007} or the Generational distance \cite{vanVeldhuizenEtAl2000} which can be used. The Hypervolume of a set of solutions with n objective variables consists of the n-dimensional volume comprised between these solutions (the approximated POF) and the optimal POF of the problem (see Figure \ref{fig:hypervolumeExample}). When the optimal POF is obtained, the hypervolume is 0. Otherwise, the higher the hypervolume, the worse the approximated POF.

On the other hand, it is also necessary to decide when the algorithm has reached a good POF and the problem must stop. For this, there exist several stopping criteria \cite{WagnerEtAl2011} in the literature. One of the stopping criteria most used consists of a comparison function which will stop the execution if the POF remains changeless for a number of generations.

\section{Description of the Multi-UAV Mission Planning Problem}\label{uavsmissionplans}
A UAV mission is typically defined as a number $n$ of \textbf{\textit{tasks}}, $T=\{t_0,t_1,...t_n\}$, performed by a team of $m$ \textbf{\textit{UAVs}}, $U=\{u_0,u_1,...u_m\}$, at a specific time interval. Each mission should be performed in a specific geographic zone. In addition, in this approach, there exist a number $l$ of \textbf{\textit{GCSs}}, $G=\{g_0,g_1,...,g_l\}$, controlling these UAVs. A solution for a mission planning problem should be the assignment of each task to a specific UAV, and each UAV to a specific GCS, ensuring that the mission can be successfully performed.

\begin{figure}[!h]
		\includegraphics[width=0.9\textwidth]{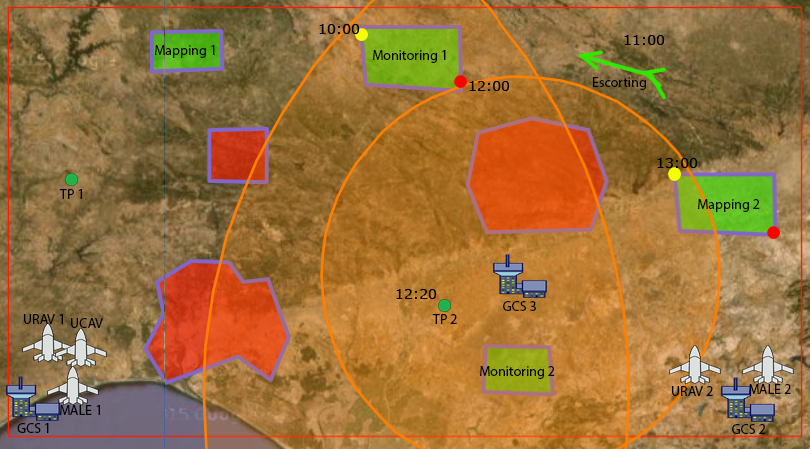}
		\centering
		\caption{Mission with 7 tasks (2 of them Multi-UAV), 5 UAVs and 3 GCSs.}
		\label{fig:bigMission}
	\end{figure}

In Figure \ref{fig:bigMission}, a Mission Scenario with 7 tasks (represented in green), 5 UAVs and 3 GCSs is presented. As can be seen in this figure, the zone of the mission could contain some No Flight Zones (NFZs), represented in red. These zones must be avoided in the trajectories of the UAVs during the mission. 

In this section, we define the different components of a mission and the computations that must be achieved to obtained the different times related to the assignments of tasks. First, we will define the types and characteristics of Tasks, UAVs and GCSs. Then, we will describe the computations that are performed in the process of task assignments.

\subsection{Task Description}
There exists different kinds of task, such as monitoring a zone or photographing a target in a specific point. These tasks are performed using the \textbf{\textit{sensors}} available by the UAVs of the mission: \textbf{EO/IR sensors}, \textbf{SAR radars}, \textbf{ISAR radars} and \textbf{MPR radars}.

\begin{definition}{}
Given a task $t \in T$, the set of sensors that can be used to perform a task is represented as $sensors(t)$.
\end{definition}

The different tasks considered in this approach and the sensor or sensors required to perform each task are represented in Table \ref{tab:tasks}.

\begin{table}[!h]
\begin{center}
\begin{tabular}{|c|C{2cm}|C{3cm}|C{1cm}|c|}
\hline
\textbf{ID} & \textbf{Name} & \textbf{Description} & \textbf{Multi-UAV} & \textbf{Sensors needed} \\
\noalign{\hrule height 2pt}
\textbf{MON} & Monitoring a zone & Fly circling in a zone during a specific time & No & \begin{minipage}{1in}
    \vskip 4pt
    \begin{itemize}
    	\item Videotracking EO/IR sensor
        \item ISAR radar
    \end{itemize}
    \vskip 4pt
\end{minipage}\\
\hline
\textbf{ES} & Escorting a path & Follow a path & No & \begin{minipage}{1.1in}
    \vskip 4pt
    \begin{itemize}
    	\item Thermal EO/IR sensor
        \item SAR radar
    \end{itemize}
    \vskip 4pt
\end{minipage}\\
\hline
\textbf{TP} & Target photographing & Go to a point and take a photo & No & \begin{minipage}{1in}
    \vskip 4pt
    \begin{itemize}
    	\item EO/IR sensor
    \end{itemize}
    \vskip 4pt
\end{minipage}\\
\hline
\textbf{MAP} & Mapping a zone & Travel a zone performing a step \& stare pattern & Yes & \begin{minipage}{1in}
    \vskip 4pt
    \begin{itemize}
    	\item SAR radar
        \item ISAR radar
        \item MPR radar
    \end{itemize}
    \vskip 4pt
\end{minipage}\\
\hline
\end{tabular}
\caption{Type of tasks. For each task, a description is given, as well as the sensors that could be used to perform it and whether this task can be performed by several UAVs or not.}
\label{tab:tasks}
\end{center}
\end{table}
    
In addition, each task has a \textit{\textbf{time interval}}, which could be specified with a \textbf{start} and \textbf{end time} for the task, or just with the \textbf{task duration}. In the last case, the start and end times will be obtained at the planning process.

On the other hand, a mission can have some \textbf{task dependencies}. There exist two types of task dependencies: \textbf{vehicle dependencies}, which impose if two tasks must be performed by the same or by different UAVs, and \textbf{time dependencies}, which constraint the relation of the time intervals of two tasks. These time dependencies are represented using \textbf{Allen's Interval Algebra}\cite{Allen1983}.

\begin{definition}{}
Given two tasks $t_1,t_2 \in T$, vehicle dependency $sameUAV(t_1,t_2)$ constraints both tasks to be performed by the same UAV.
\end{definition}

\begin{definition}{}
Given two tasks $t_1,t_2 \in T$, vehicle dependency $diffUAV(t_1,t_2)$ constraints both tasks to be performed by different UAVs.
\end{definition}

\subsection{UAV Description}
The UAVs of a mission, $u \in U$, have some \textbf{\textit{features}} that must be considered when checking if a plan is correct. These features are presented in Table \ref{tab:uav_features}.

\begin{table}
\begin{center}
\begin{tabular}{|C{1cm}|C{2.5cm}|C{4.5cm}|c|}
\hline
\multicolumn{2}{|c|}{\textbf{Feature}} & \textbf{Description} & \textbf{Symbol} \\
\noalign{\hrule height 2pt}
\multicolumn{2}{|c|}{\textbf{Initial Position}} & The position of the UAV at the beginning of the mission & $pos_u$ \\
\hline
\multicolumn{2}{|c|}{\textbf{Initial Fuel}} & The fuel of the UAV at the beginning of the mission & $fuel(u)$ \\
\hline
\multicolumn{2}{|c|}{\textbf{Available sensors}} & The sensors contained in the UAV & $sensors(u)$ \\
\hline
\multicolumn{2}{|c|}{\textbf{Range}} & The maximum distance that the UAV can traverse in the mission & $range(u)$ \\
\hline
\multicolumn{2}{|c|}{\textbf{Autonomy}} & The maximum time that the UAV can stay in fly & $autonomy(u)$ \\
\hline
\multicolumn{2}{|c|}{\textbf{Cost}} & The cost per hour of use of the UAV & $cost(u)$ \\
\hline
\multicolumn{2}{|c|}{\textbf{Max. Speed}} & The maximum speed attainable by the UAV & $max_{speed}(u)$ \\
\hline
\multicolumn{2}{|c|}{\textbf{Max. Altitude}} & The higher altitude that the UAV can reach  & $max_{alt}(u)$ \\
\hline
\multicolumn{2}{|c|}{\textbf{Max. Fuel}} & The maximum fuel capacity of the UAV tank  & $max_{fuel}(u)$ \\
\hline
\multicolumn{2}{|c|}{\textbf{Flight profiles (FP)}} & One or more profiles that specify at each moment the fly features of the UAV & $fps(u)$ \\
\hline
FP & \textbf{Speed} & Speed of the UAV for a flight profile & $speed(fp_u)$ \\
\hline
FP & \textbf{Fuel Consumption Ratio} & Fuel consumption by hour of the UAV for a flight profile & $fuelRatio(fp_u)$ \\
\hline
FP & \textbf{Altitude} & Altitude of the UAV when using a route flight profile & $altitude(fp_u)$ \\
\hline
FP & \textbf{Angle} & Angle of the UAV when using a climb/descent flight profile & $angle(fp_u)$ \\
\hline
\end{tabular}
\caption{Different UAV features considered.}
\label{tab:uav_features}
\end{center}
\end{table}

On the other hand, during the mission, the UAV will be positioned in different points at each moment.

\begin{definition}{}
Given a UAV $u \in U$, the position of $u$ at any time $t \in \mathbb{R}$ is represented as $pos(u,t)$.
\end{definition}

Each type of UAV, $type(u)$, will have different values for these features. In this approach, four basic \textbf{\textit{types}} of UAVs have been considered. These are described in Table \ref{tab:uavs}.

\begin{table}
\begin{center}
\scalebox{0.82}{
\begin{tabular}{|c|C{1cm}|C{1.6cm}|C{1.1cm}|C{1cm}|C{1.4cm}|C{1cm}|c|}
\hline
\textbf{Name} & \textbf{Range (NM)} & \textbf{Autonomy (h)} & \textbf{Cost/h} & \textbf{Max. Speed (kt)} & \textbf{Max. Altitude (ft)} & \textbf{Max. Fuel (kg)} & \textbf{Available Sensors} \\
\noalign{\hrule height 2pt}
\textbf{URAV} & 1000 & 20 & 5 & 120 & 20000 & 500 & \begin{minipage}{1in}
    \vskip 4pt
    \begin{itemize}
    	\item Videotracking and thermal EO/IR sensor
    \end{itemize}
    \vskip 4pt
\end{minipage}\\
\hline
\textbf{MALE}  & 5000 & 30 & 10 & 250 & 40000 & 2500 & \begin{minipage}{1.1in}
    \vskip 4pt
    \begin{itemize}
    	\item EO/IR sensor
        \item MPR radar
    \end{itemize}
    \vskip 4pt
\end{minipage}\\
\hline
\textbf{HALE}  & 15000 & 40 & 15 & 400 & 65000 & 6000 & \begin{minipage}{1in}
    \vskip 4pt
    \begin{itemize}
    	\item Videotracking EO/IR sensor
        \item ISAR radar
    \end{itemize}
    \vskip 4pt
\end{minipage}\\
\hline
\textbf{UCAV}  & 1500 & 15 & 25 & 450 & 35000 & 9000 & \begin{minipage}{1in}
    \vskip 4pt
    \begin{itemize}
    	\item EO/IR sensor
        \item SAR radar
    \end{itemize}
    \vskip 4pt
\end{minipage}\\
\hline
\end{tabular}
}
\caption{Different types of UAVs considered and their features.}
\label{tab:uavs}
\end{center}
\end{table}

\subsection{GCS Description}
To solve Multi-UAV missions, it is necessary to use several GCSs controlling the UAVs. Therefore, the problem is Multi-GCS, and it should be checked that each UAV is controlled by an appropriate GCS. Every GCS $g \in G$ has some features to be considered that are represented in Table \ref{tab:gcs_features}.

\begin{table}[h]
\begin{center}
\begin{tabular}{|c|C{4.2cm}|c|}
\hline
\textbf{Feature} & \textbf{Description} & \textbf{Symbol} \\
\noalign{\hrule height 2pt}
\textbf{Position} & The position of the GCS & $pos_g$ \\
\hline
\textbf{Max. Number of UAVs} & The maximum number of UAVs that the GCS can control & $maxNum(g)$ \\
\hline
\textbf{Permitted Types} & The permitted types of UAVs that the GCS can control & $types(g)$ \\
\hline
\textbf{Coverage} & The within range of the GCS & $coverage(g)$ \\
\hline
\end{tabular}
\caption{Basic features considered for a GCS.}
\label{tab:gcs_features}
\end{center}
\end{table}

In Figure \ref{fig:bigMission}, the coverage is represented for each GCS in translucent orange. It can be appreciated that GCS3 has a low coverage, while GCS1 and GCS2 have a higher range.

\subsection{Task Assignment Processes}\label{taskassignment}
\begin{figure}[h]
		\includegraphics[width=\textwidth]{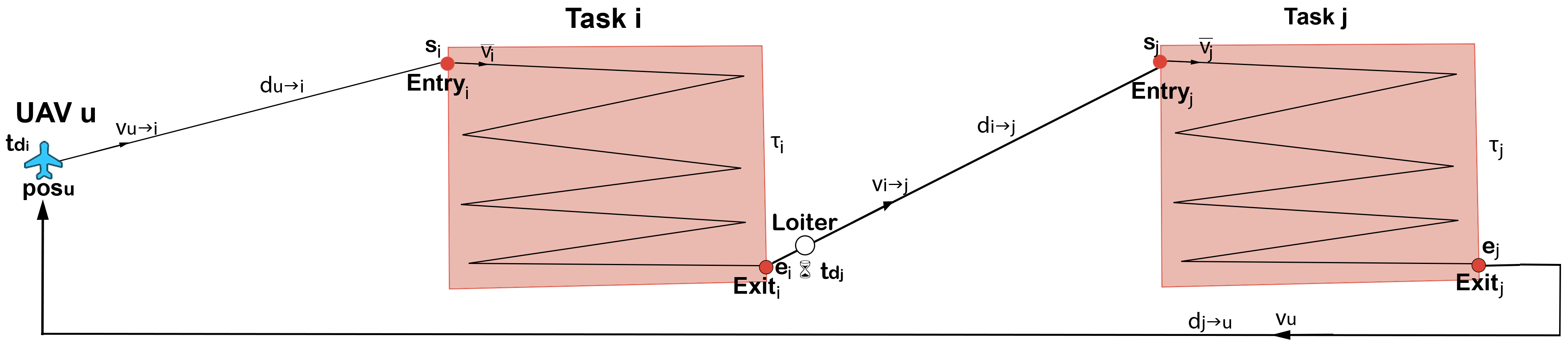}
		\centering
		\caption{Example of assignment of a UAV to two tasks. The path to each task, the task performance, the loiter and the return phases are represented, as well as every time point and duration related.}
		\label{fig:taskexample}
	\end{figure}

In figure \ref{fig:taskexample}, an assignment of a UAV $u$ to two tasks $i$ and $j$ is represented, and it can be seen the different times computed in the process. In this assignment process, it is necessary to compute several variables related to time, fuel consumption and distance traversed, in order to validate that the task can be fulfilled at its time interval using the assigned UAV.

The variables related to \textbf{time} that must be computed in this task assignment process are:

\begin{itemize}

\item The \textbf{departure time} when the vehicle starts moving to the task zone. In Figure \ref{fig:taskexample}, it is represented as ${t_d}_i$ for task $i$, and ${t_d}_j$ for task $j$.

\item The \textbf{duration of the path} between the departure of the UAV and the start of the task. In order to compute this duration, the \textbf{path flight profile} used by the UAV in this path must be set. With the \textbf{speed} ($v_i$) provided by this profile and the distance from the UAV departure position to the task zone, it is possible to compute the duration of the path ($d_{u \to i}$).

\item The \textbf{start time} of the task. This time could be fixed in the definition of the task. If not, it is computed during the assignment process. It is represented as $s_i$ in Figure \ref{fig:taskexample} for task $i$.

\item The \textbf{duration of the task} ($\tau_i$). This time could be given (e.g. in monitoring tasks) or must be computed (e.g. in mapping tasks). In the second case, it is necessary to know the speed ($\overline{v_i}$) of the UAV in the task performance. This is given by the \textbf{sensor} used by the UAV to perform the task, which provides the \textbf{optimum speed and altitude} for its use.

\item The \textbf{end time} of the task. This time could be fixed in the definition of the task. If not, it is computed during the assignment process. It is represented as $e_i$ in Figure \ref{fig:taskexample} for task $i$.

\item The \textbf{duration of the loiter}. When start and end times of tasks are fixed, it may happen that the time when a UAV finishes a task does not meet the time when the UAV departs for the next task. The difference between these two times is known as the \textbf{loiter duration} for the second task.

\item The \textbf{duration of the return}. In order to compute this duration, the \textbf{return flight profile} used by the UAV in this return must be set. With the speed ($v_u$) provided by this profile and the distance from the zone of the last task of the UAV to its initial position, it is possible to compute the duration of the return ($d_{j \to u}$).

\item The \textbf{return time} when the UAV has returned to its initial position. It is computed as the sum of the end time of the last task performed by the UAV and the duration of the return.

\end{itemize}

On the other hand, there are some variables related to fuel consumption that must be computed in this task assignment process. These variables are computed using the previous durations and the fuel consumption ratio given by the flight profile used in each case. Specifically, these variables are: The \textbf{fuel consumption of the path}; The \textbf{fuel consumption of the task}; The \textbf{fuel consumption of the loiter}; and the \textbf{fuel consumption of the return}.

Finally, the variables related to the distance traversed are computed as the sums of distances between the points of the path employed in each case. These variables are: The \textbf{distance of the path}; The \textbf{distance of the task}; The \textbf{distance of the loiter}; and the \textbf{distance of the return}.

\begin{definition}{}
Given two points $p_1,p_2 \in \mathbb{R}^3$ in 3D geographical coordinates (longitude, latitude, altitude), we define the distance function $distance(p_1,p_2)$ between them as the 3D distance in WGS84 system.
\end{definition}

\section{Modelling the MPP as a CSP}\label{cspmodelling}

In this section, we define how the MPP can be modelled using a CSP. First, we define which are the variables of the CSP and their domain. Then, we explain the different constraints considered for the MPP.

\subsection{CSP variables}
Looking at the assumptions explained so far in the previous section, the variables of the CSP that we have considered are as follows:

\begin{itemize}
	\item \textbf{Assignments ($assign$) of tasks to UAVs}. As some tasks could be Multi-UAV, these variables are represented as a binary array of size $n \times m$. An assignment $assign[t,u]=1$ means that task $t$ is assigned to UAV $u$.
	\item \textbf{Orders ($order$)}, which define the order in which each UAV performs the tasks assigned to it. These variables are necessary when start and end times of tasks are not fixed, and they are represented as an array of size $n \times m$. Their domain is $[-1 .. n-1]$, where $-1$ is only assigned when the UAV does not perform the task.
    \item \textbf{Assignments of UAVs to GCSs ($gcss$)}. There are $m$ variables of this type, and their domain is $[-1 .. l-1]$, where $-1$ is only assigned when the UAV is not assigned to any task.
	\item \textbf{Path Flight Profiles ($fpPath$)}, setting the flight profile that the vehicle must take for the path performance. These variables are represented as a $n \times m$ array, and their domain are the flight profiles of the UAV in the column: $fpPath[t,u] \in fps(u)$.
	\item \textbf{Return Flight Profiles ($fpReturn$)}, similar to the previous set of variables but for the return path of each UAV. There are $m$ variables of this type, and their domain is the same as the previous variables: $fpReturn[u] \in fps(u)$.
	\item \textbf{Sensor used in the task performance ($sensTask$)}. These variables set the sensor of the vehicle that will be used during the task performance. It will be necessary to consider these variables just in the case that the vehicle performing the task has several sensors that could perform that task. These variables are represented as a $n \times m$ array, and their domain are the sensors of the task and UAV available for that assignment: $sensTask[t,u] \in sensors(t) \cap sensors(u)$.
\end{itemize}

On the other hand, there are some extra variables that will be computed during the propagation phase of the CSP. These variables are directly related to the variables computed in section \ref{taskassignment}: $departure$, $durPath$, $start$, $durTask$, $end$, $durLoiter$, $durReturn$, $returnTime$, $fuelPath$, $fuelTask$, $fuelLoiter$, $fuelReturn$, $distancePath$, $distanceTask$, $distanceLoiter$ and $distanceReturn$.

\subsection{CSP constraints}\label{constraints}
Now, we define the different constraints of the CSP, which consider all the specifications explained so far:

\begin{enumerate}

\item \textbf{Sensor constraints}: they check if a UAV has the sensor needed to perform its assigned tasks. Let $sensors(u)$ denote the sensors available for UAV $u$ and $sensors(t)$ the sensors that could perform the task $t$ then:

\begin{equation}
  \forall t \in T \quad \forall u \in U \quad assign[t,u]=1 \Rightarrow |sensors(t) \cap sensors(u)| > 0
\end{equation}

\item \textbf{Order constraints}: they assure that the values of the order variables are less than the number of tasks assigned to the UAV performing that task:

\begin{align}
\forall t \in T \quad \forall u \in U \quad & assign[t,u]=1 \Rightarrow \nonumber \\
& \qquad order[t,u] < \sharp \left\{ { \tau \in T }|{ assign[\tau,u]=1 } \right\}
\end{align}

and if two tasks are assigned to the same UAV, they have different orders:

\begin{align}
\forall i,j \in T \quad \forall u \in U \quad assign[i,u]&=assign[j,u]=1 \Rightarrow \nonumber \\
& \qquad order[i,u]\neq order[j,u]
\end{align}

\item \textbf{GCS constraints}: they assure that the GCSs assignments are correct. First, it is necessary to assure that the UAVs assigned to the GCS are of a type supported by that GCS (both in initial assignment and during tasks performance):

\begin{equation}
\forall u \in U \quad \forall g \in G \quad gcss[u]=g \Rightarrow type(u) \subset types(g)
\end{equation}

Then, a constraint assures that the maximum number of UAVs that a GCS can handle is not overpassed at any moment:

\begin{equation}
\forall g \in G \quad \sharp \left\{ { u\in U }|{ gcss[u]=g } \right\} < maxNum(g)
\end{equation}

Finally, it is necessary to check that GCS can coverage the UAV during the mission:

\begin{align}
\forall u \in U \quad \forall g \in G \quad & gcss[u]=g \Rightarrow \nonumber \\
& \forall t \in \mathbb{R} \quad distance(pos(u,t),pos_g) \leq coverage(g)
\end{align}

\item \textbf{Temporal constraints}: they assure the consistency of all the time variables considered. First, it is necessary to assure that the start time of the task equals the sum of the departure time and the duration for the path:

\begin{align}
\forall t \in T \quad \forall u \in U \quad & assign[t,u]=1 \Rightarrow \nonumber \\ 
& \qquad departure[t,u]+durPath[t,u]=start[t,u]
\end{align}

and that end time is the sum of the start time and the duration of the task:

\begin{equation}
\forall t \in T \quad \forall u \in U \quad assign[t,u]=1 \Rightarrow start[t,u]+durTask[t,u]=end[t,u]
\end{equation}

Then, the duration of the path is computed as the distance traversed in the path divided by the speed given by the path flight profile:

\begin{equation}
\forall t \in T \quad \forall u \in U \quad assign[t,u]=1 \Rightarrow durPath[t,u]=\frac{distancePath[t,u]}{speed(fpPath[t,u])}
\end{equation}

If tasks have fixed start and end times, then it is necessary to compute the duration of the loiter as the difference between the end of a task and the departure for its consecutive task:

\begin{align}
\forall i,j \in T \quad \forall u \in U \quad & assign[i,u]=assign[j,u]=1 \nonumber \\
& \wedge order[i,u]=order[j,u]-1 \nonumber \\
& \Rightarrow durLoiter[j,u]=departure[j,u]-end[i,u]
\end{align}

On the other hand, the duration of the return is computed as the distance traversed in the return path divided by the speed given by the return flight profile:

\begin{equation}
\forall u \in U \quad durReturn[u]=\frac{distanceReturn[u]}{speed(fpReturn[u])}
\end{equation}

Once we have computed the return path duration, we can compute the return time as the sum of the end of the last task performed by the UAV and this return duration:

\begin{align}
\forall t \in T \quad \forall u \in U \quad & assign[t,u]=1 \nonumber \\
& \wedge order[t,u]=\sharp \left\{ { \tau\in T }|{ assign[\tau,u]=1 } \right\}-1 \nonumber \\
& \Rightarrow returnTime[u]=end[t,u]-durReturn[u]
\end{align}

Finally, it is necessary to assure that two tasks that collide in time are never assigned to the same UAV:

\begin{align}
\forall i,j \in T & \quad \forall u \in U \quad  assign[i,u] = assign[j,u]=1 \nonumber \\
& \wedge  order[i,u] <  order[j,u] \Rightarrow  end[i,u] \leq departure[j,u]
\end{align}

\item \textbf{Dependency Constraints}: these constraints are related to the time and vehicle dependencies mentioned before. The time dependency constraints, based on Allen's Interval Algebra \cite{Allen1983}, for each pair of tasks $i$ and $j$, assuming $\forall i,j \in T \quad \forall u \in U \quad assign[i,u] = assign[j,u]=1$, are as follow:

\begin{equation}
i < j \Rightarrow \quad end[i,u] \le start[j,u]
\end{equation}
\begin{equation}
i\quad m\quad j \Rightarrow end[i,u]=start[j,u]
\end{equation}
\begin{equation}
i\quad o\quad j \Rightarrow \begin{cases} start[i,u] \le start[j,u] \\ end[i,u] \ge start[j,u] \\ end[i,u] \le end[j,u] \end{cases}
\end{equation}
\begin{equation}
i\quad s\quad j \Rightarrow \begin{cases} start[i,u]=start[j,u] \\ end[i,u] \le end[j,u] \end{cases}
\end{equation}
\begin{equation}
i\quad d\quad j \Rightarrow \begin{cases} start[i,u] \ge start[j,u] \\ end[i,u] \le end[j,u] \end{cases}
\end{equation}
\begin{equation}
i\quad f\quad j \Rightarrow \begin{cases} start[i,u] \ge start[j,u] \\ end[i,u]=end[j,u] \end{cases}
\end{equation}
\begin{equation}
i = j \Rightarrow \begin{cases} start[i,u]=start[j,u] \\ end[i,u]=end[j,u] \end{cases}
\end{equation}

On the other hand, vehicle dependencies imply the following constraints:

\begin{equation}
\forall i,j \in T \quad sameUAV(i,j) \Rightarrow \forall u \in U \quad assign[i,u]=assign[j,u]
\end{equation}
\begin{equation}
\forall i,j \in T \quad diffUAV(i,j) \Rightarrow \forall u \in U \quad assign[i,u]\neq assign[j,u]
\end{equation}

\item \textbf{Autonomy constraints}: they assure that the total flight time for each vehicle is less than its autonomy:

\begin{align}
\forall u \in U \quad flightTime[u] &= \sum _{\mathclap{\substack{t\in T\\ assign[t]=u}}}{ (durPath[t]+durTask[t]+durLoiter[u]) } \nonumber \\
& \qquad \qquad + durReturn[u] < autonomy(u)
\end{align}

\item \textbf{Distance constraints}: they assure that the distance traversed by each vehicle is less than its range:

\begin{align}
\forall u \in U \quad & distance[u] = \sum _{\mathclap{\substack{t\in T\\ assign[t]=u}}}{ (distancePath[t]+distanceTask[t] } \nonumber \\
& \quad +distanceLoiter[u]) + distanceReturn[u] < range(u)
\end{align}

To compute these distances, we have used GeographicLb\footnote{http://geographiclib.sourceforge.net/} for the computation of distance and points in Geographic coordinates; and Theta* \cite{NashEtAl2007} to perform a path between these points avoiding No Flight Zones (NFZ) and terrain obstacles. The elevation of the terrain has been read from DTED maps using GDAL\footnote{http://www.gdal.org/}.

\item \textbf{Fuel constraints}: they assure that the fuel consumed by each vehicle is less than its initial fuel ${fuel}_u$:

\begin{align}
\forall u \in U fuel[u]=\sum _{\mathclap{\substack{t\in T\\ assign[t]=u}}}{ (fuelPath[t]+fuelTask[t]) }+ fuelReturn[u] \nonumber  \\
< fuel(u)
\end{align}

Each one of these fuel consumptions is computed as the product of its associated duration and fuel consumption ratio. For example, the fuel consumption for the path is computed multiplying the fuel consumption ratio given by the path flight profile and the duration of the path:

\begin{align}
\forall t \in T \quad & \forall u \in U \quad assign[t,u]=1 \Rightarrow  \nonumber \\ 
& fuelPath[t,u]=durPath[t,u] \times fuelRatio(fpPath[t,u])
\end{align}

\end{enumerate}

\section{MOGA-CSP Algorithm for Multi-UAV Mission Planning Problems}\label{mogaapproach}
Given the big amount of solutions that the problem can generate and the huge amount of constraints involved in the search of solutions, we have decided to use a hybrid approach based on MOGAs and CSPs to solve MPPs. In this new approach, the constraints of the problem have been applied as penalty function in the evaluation phase of the genetic algorithm. This section describes this algorithm, including the encoding, the fitness function designed, and the genetic operators implemented.

\subsection{Encoding}
\begin{figure}[!h]
		\includegraphics[width=\textwidth]{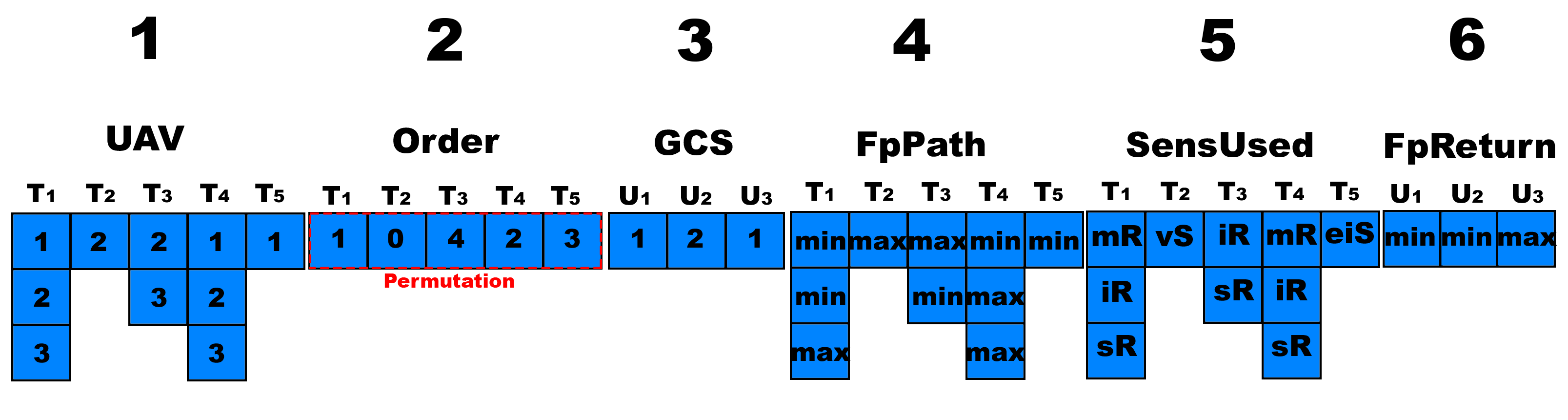}
		\centering
		\caption{Example of an individivual that represents a possible solution for a problem with 5 tasks, 3 UAVs and 2 GCSs.}
		\label{fig:chromosome}
	\end{figure}

To encode the Multi-UAV MPP, a representation based on six different alleles have been designed (see Figure \ref{fig:chromosome}). Each allele is used to encode the features that have been described in previous sections representing a complete solution that will be optimized by MOGA algorithm. Next, a short description for each allele is given:

\begin{enumerate}
	\item \textbf{UAVs assigned to each task}. If the $T_i$ task is Multi-UAV, then this cell contains a vector representing the different UAVs assigned to this task. 
	\item \textbf{Permutation of the task orders}. These values indicate the absolute order of the tasks. It is only used if there are several tasks assigned to the same UAV and some of them do not have the start and end times fixed.
	\item \textbf{GCSs controlling each UAV}. 
	\item \textbf{Flight profiles used for each UAV to each assigned task}. As in the first allele, some of the cells could contain a vector if the corresponding task is performed by several UAVs. 
	\item \textbf{Sensors used for the task performance} by each UAV.
	\item \textbf{Flight Profiles used by each UAV to return} to the base.
\end{enumerate}

An example of this representation is shown in Figure \ref{fig:chromosome}. Firstly, this figure shows a mission with 5 tasks. Assuming that there is not any task with start and end time fixed, it is necessary to use the permutation allele for the task orders (2). Using together, this allele and the allele of UAV assignments (1), we have that UAV 1 performs tasks 1, 4 and 5 in this order; UAV 2 performs tasks 2, 1, 4 and 3; and UAV 3 performs tasks 1, 4 and 3. On the other hand, according to allele of GCCs information (3), we have that UAVs 1 and 3 are controlled by GCS 1, while UAV2 is controlled by GCS 2. Furthermore, in the allele of Flight profiles per task (4), we can see that UAV 1 uses minimum consumption flight profile for all its assigned tasks; UAV 2 uses minimum consumption profile for task 1, and maximum speed profile for the rest of tasks, and UAV 3 uses minimum consumption profile for task 3, while maximum speed profile for the rest of tasks. Regarding the sensors used (5), it can be seen that task 1 is performed by UAV 1 using MPR radar (mR) sensor, while UAV 2 uses an ISAR radar (iR), and UAV 3 uses a SAR radar (sR); task 2 is performed using EO/IR sensor (eiS), etc. Finally, the last allele (6) represents that UAVs 1 and 2 use minimum consumption profile for their return path, while UAV 3 uses maximum speed profile.

A key point in this representation is that only a valid sensor to perform the task assigned could be used for the allele of sensors used per task. With this, the algorithm is avoiding some invalid solutions due to sensor constraints.

\subsection{Fitness Function}\label{fitness}
Evaluation is computed in terms of a fitness function composed by two check steps. First, for a given solution, it handles that all constraints are fulfilled. If not, it acts as a penalty function, giving the solution the worst possible value so it would not be evolved in future generations. If all constraints are fulfilled, the fitness function works as a multi-objective function minimizing the objectives of the problem. For this purpose, we have considered the optimization variables described in Table \ref{tab:fitness}.

\begin{table}[h]
\begin{center}
\begin{tabular}{|c||C{4cm}|c|}
\hline
\textbf{Variable} & \textbf{Description} & \textbf{Formula} \\
\noalign{\hrule height 2pt}
$N_{uavs}$ & The number of UAVs used in the mission & $\sharp \left\{ { u\in U }|{ \exists t\in T \quad assign[t]=u } \right\}$\\
\hline
$totalFlightTime$ & The total flight time of all the UAVs during the mission & $\sum _{ u\in U }{ flightTime[u] }$\\
\hline
$totalFuel$ & The total fuel consumed by all the UAVs during the mission & $\sum _{ u\in U }{ fuelConsumed[u] }$\\
\hline
$totalDistance$ & The total distance traversed by all the UAVs in the mission & $\sum _{ u\in U }{ distance[u] }$\\
\hline
$totalCost$ & The total cost of the mission, computed as the sum of the individual cost of each UAV & $\sum _{ u\in U }{ cost(u) \times flightTime[u] }$\\
\hline
$makespan$ & The time when the mission ends (all UAVs have returned) & $\max _{ u\in U }{ returnTime[u] }$\\
\hline
\end{tabular}
\caption{Optimization variables used in the fitness function by MOGA-CSP algorithm.}
\label{tab:fitness}
\end{center}
\end{table}

The multi-objective fitness function compares the solution evaluated with the stored solutions in order to obtain the Pareto-Optimality Frontier (POF) based on the NSGA-II approach \cite{DebEtAl2002}.

\subsection{Algorithm}
In this new approach, as can be seen in Algorithm \ref{alg:moga}, after evaluating the individuals of the population with the fitness previously explained (Line 8), a \textit{N} elitist selection is performed. It means, that a number $N$ of best individuals in the population is retained (Line 9). Then, a roulette wheel selection over these $N$ individuals (Line 12) selects those that will be applied the genetic operators.

\begin{algorithm}[h]
 \caption{Hybrid MOGA-CSP algorithm for Mission Planning Problems}
 \label{alg:moga}
\DontPrintSemicolon
\KwIn{ A mission $M=(T,U,G)$ where $T$ is a set of tasks to perform denoted by $\{t_1, \dots, t_n\}$, $U$ is a set of UAVs denoted by $\{u_1, \dots, u_m\}$ and $G$ is a set of GCSs denoted by $\{g_1, \dots, g_l\}$. And positive numbers $generations$, $population$, $\mu$, $\lambda$, $mutprobability$ and $stopGeneration$}
\KwOut{POF obtained with best solutions}
  $S \gets$ randomly generated set of $population$ of $p$ chromosomes with 6 alleles representing the tasks assignments to UAVs, the orders, the UAVs assignments to GCSs, the path flight profiles, the sensors used and the return flight profiles \;
 $i \gets 1$ \;
 $convergence \gets 0$ \;
 $pof \gets createPOF(S)$ \;
 \While{$i \leq generations \land convergence < stopGenerations$}{
  	$F \gets \emptyset$ \;
  	\For{$j \gets 1$ \textbf{to} $p$}{
    	$F \gets Fitness(S_j)$ \;
    }
    $Sbest \gets SelectNBest(\mu,F)$ \;
    $newS \gets Sbest$ \;
    \For{$j \gets \mu$ \textbf{to} $\lambda$}{
      $p1,p2 \gets RouleWheelSelection(Sbest$) \;
      $i1,i2 \gets Crossover(p1, p2)$ \;
      $i1 \gets Mutation(i1,mutprobability)$ \;
      $i2 \gets Mutation(i2,mutprobability)$ \;
      $newS \gets newS \cup \{i1, i2\}$ \;
    }
    $NSGA2_{UpdatePopulation}(S,newS)$ \;
    $newpof \gets createPOF(S)$ \;
    \If {$newpof = pof$}{
    	$convergence \gets convergence + 1$ \;
    }
    $pof = newpof$ \;
    $i \gets i + 1$ \;
 }
 \Return{pof}\;
\end{algorithm}

\begin{figure}[!h]
		\includegraphics[width=\textwidth]{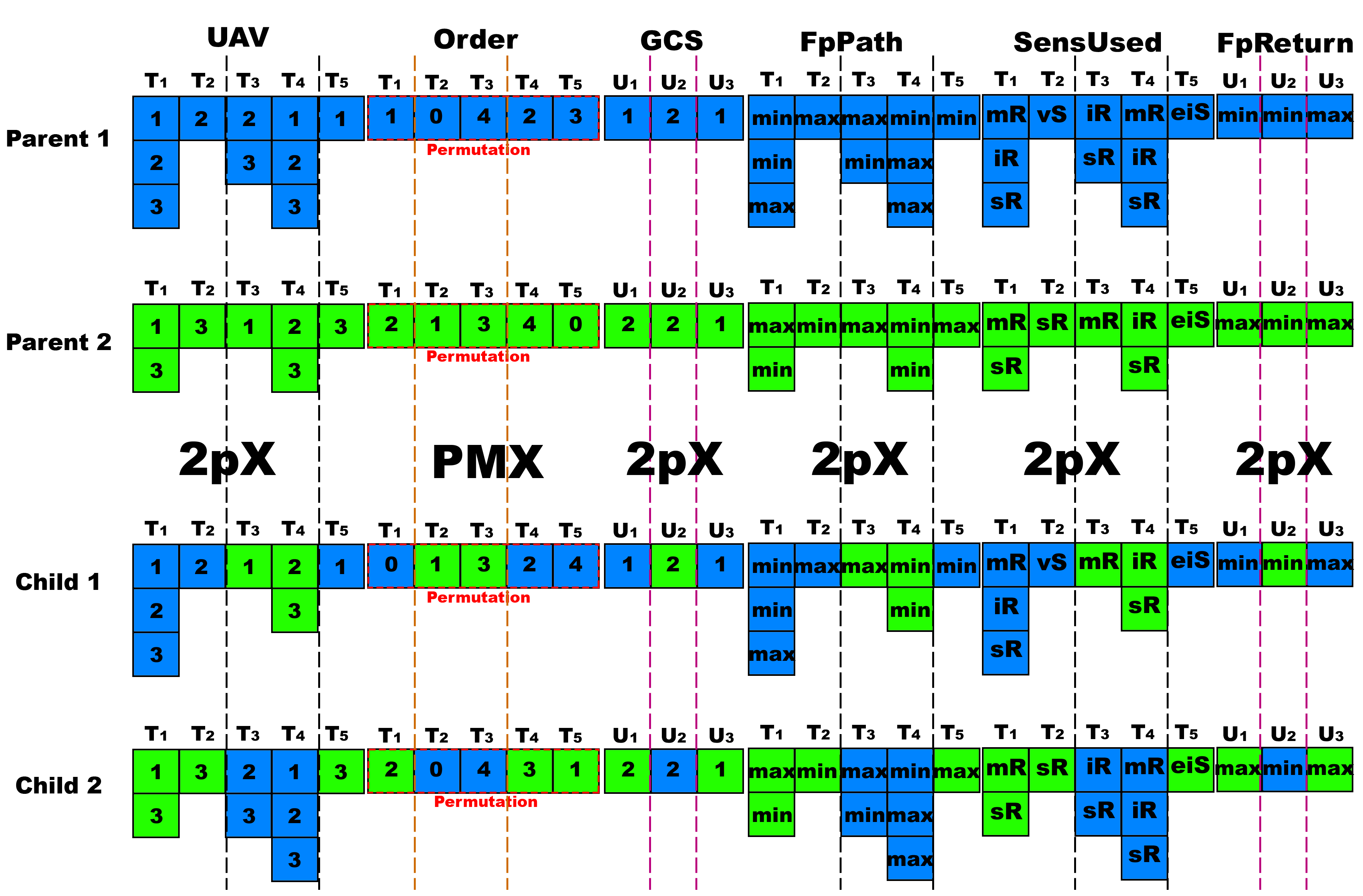}
		\centering
		\caption{Example of crossover of two parents with 5 tasks, 3 UAVs and 2 GCSs. Each allele is performed a different type of crossover: UAV, FpPath and SensUsed are performed a 2-point crossover; Order is applied a PMX crossover, and GCS and FpReturn are applied another 2-point crossover.}
		\label{fig:crossover}
	\end{figure}

Next, we use a proper crossover operator (Line 13) to combine the chromosomes of each pair of parents to generate a new pair of children. This operator consists of a specific crossover operation for each of the alleles of the representation. The first allele performs a 2-point crossover, and the same cross points used for this allele are reused for the fourth and fifth allele in order to maintain the size for Multi-UAV tasks and the consistency of the sensors used. On the other hand, in the second allele, as it is a permutation, is applied a Partially-Matched Crossover (PMX). This passes a swatch of values from one parent to the other and then performs a replacement of the invalid values of the new child based on its previous parent. Finally, in the third and sixth alleles are applied another 2-point crossover (with different points than the previous). Figure \ref{fig:crossover} shows an example of this crossover operation, where the first, fourth and fifth allele have selected points 2 and 4 for the 2-point crossover. In the second allele a swatch composed of task $T_2 .. T_3$ has been selected for the PMX crossover, and finally, the third and sixth allele have selected points 1 and 2 for the 2-point crossover.

\begin{figure}[!h]
		\includegraphics[width=\textwidth]{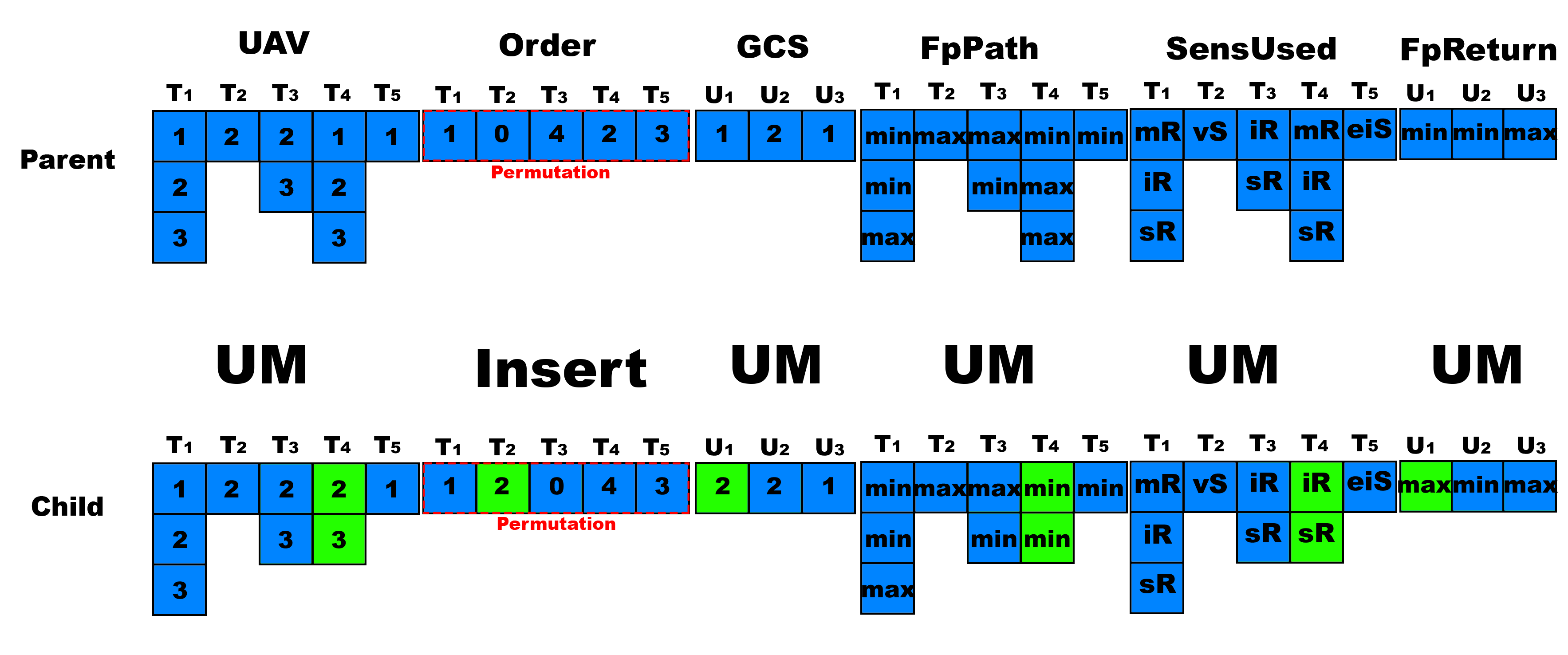}
		\centering
		\caption{Example of Mutation for an individual with 5 tasks, 3 UAVs and 2 GCSs. Each allele is performed a different type of mutation: UAV, FpPath and SensUsed are performed an uniform; Order is applied an Insert mutation, and GCS and FpReturn are applied another uniform mutation.}
		\label{fig:mutation}
	\end{figure}

Once the new pair of individuals has been generated from crossover operation, a mutation operator (Line 14) will be applied to them depending on a probability $P_m$ (usually low, $\sim 5\%$). This genetic operator helps to avoid that the obtained solutions stagnate at local minimums. This mutation operator is designed to perform an uniform mutation over the same genes for the first, fourth and fifth allele in order to maintain the size of Multi-UAV tasks and avoid invalid solutions accomplishing sensor constraints. On the other hand, the second allele is applied an Insert Mutation, which will select two random positions from the permutation and move the second one next to the first one. Finally, the third and sixth allele are performed another uniform mutation. Figure \ref{fig:mutation} presents an example of this mutation, where $T_4$ has been mutated for the first, fourth and fifth allele, the insert mutation has moved the value of $T_4$ next to $T_1$, and the third and sixth allele have mutated the value of $U_1$.

Finally, after the population is updated by NSGA-II (Line 16), the stopping criteria designed for this algorithm compares the POF obtained so far in each generation with the POF from the previous generation (Line 18). If this POF remains unchangeable for a number of generations, then the algorithm will stop and return this POF.

\section{Experiments}\label{experiments}
In this section we explain the experiments carried out to test the functionality of the new MOGA-CSP approach for MPP. For this purpose, we have designed several missions with different configurations of tasks, UAVs, GCSs and NFZs in order to check the different characteristics of the model. These datasets are described in Table \ref{tab:datasets} which shows the characteristics of the model that are checked for each one.

The first experiment shows the results obtained when the different objectives are optimized individually and by pairs, and compare it with their optimization all together. From this experiment, we will obtain which variables are the most appropriate to use for this problem for the MOGA-CSP.

Finally, all datasets are tested using the objective variables obtained in the previous experiment. In order to evaluate the performance of the algorithm the Hypervolume metric is calculated. To apply this metric, it is necessary to compute the optimal POF using the MOBB algorithm for each dataset. For this purpose MOBB algorithm, provided from Rodriguez-Fernandez et. al. approach \cite{Rodriguez-FernandezEtAl2015a} is applied. Then, the solutions returned by the MOGA-CSP are compared with the MOBB results to analyse their optimality.

\begin{table}[h]
\centering
\scalebox{0.80}{
\begin{tabular}{|c|c||c|c|c|c|c|c|c| }
  \hline
  \multicolumn{2}{|c|}{Dataset} & \multirow{2}{*}{Tasks} & \multirow{2}{*}{UAVs} & \multirow{2}{*}{GCSs} & \multirow{2}{*}{NFZs} & \multicolumn{3}{|c|}{Task Times} \\
  \cline{1-2}
  \cline{7-9}
   Id. & Description &  &  &  &  & Fixed & Unfixed & Deps. \\
  \noalign{\hrule height 2pt}	
  1& \begin{minipage}{3.0cm}
    \vskip 4pt
    Simple mission with fixed times
    \vskip 4pt
\end{minipage} & \begin{minipage}{0.6in}
    \vskip 4pt
    \begin{itemize}
    	\item 2 MON
        \item 2 ES
        \item 2 TP
    \end{itemize}
    \vskip 4pt
\end{minipage} & \begin{minipage}{0.7in}
    \vskip 4pt
    \begin{itemize}
    	\item 1 HALE
        \item 1 MALE
        \item 1 UCAV
        \item 1 URAV
    \end{itemize}
    \vskip 4pt
\end{minipage} & 1 & 0 & 6 & 0 & 0 \\
  \hline
  2 & \begin{minipage}{3.0cm}
    \vskip 4pt
    Path avoidance
    \vskip 4pt
\end{minipage} & \begin{minipage}{0.6in}
    \vskip 4pt
    \begin{itemize}
    	\item 2 MON
        \item 2 ES
        \item 2 TP
    \end{itemize}
    \vskip 4pt
\end{minipage} & \begin{minipage}{0.7in}
    \vskip 4pt
    \begin{itemize}
    	\item 1 HALE
        \item 1 MALE
        \item 1 UCAV
        \item 1 URAV
    \end{itemize}
    \vskip 4pt
\end{minipage} & 1 & 1 & 6 & 0 & 0 \\
  \hline
  3 &  \begin{minipage}{3.0cm}
    \vskip 4pt
    Multi-UAV tasks
    \vskip 4pt
\end{minipage} & \begin{minipage}{0.6in}
    \vskip 4pt
    \begin{itemize}
    	\item 3 MAP
    \end{itemize}
    \vskip 4pt
\end{minipage} & \begin{minipage}{0.7in}
    \vskip 4pt
    \begin{itemize}
    	\item 1 HALE
        \item 1 MALE
    \end{itemize}
    \vskip 4pt
\end{minipage} & 1 & 0 & 0 & 3 & 0 \\
  \hline
  4a & \begin{minipage}{3.1cm}
    \vskip 4pt
    Multi-GCS with fixed times
    \vskip 4pt
\end{minipage} & \begin{minipage}{0.6in}
    \vskip 4pt
    \begin{itemize}
    	\item 2 MON
        \item 2 ES
        \item 2 TP
    \end{itemize}
    \vskip 4pt
\end{minipage} & \begin{minipage}{0.7in}
    \vskip 4pt
    \begin{itemize}
    	\item 1 HALE
        \item 1 MALE
        \item 1 UCAV
        \item 2 URAV
    \end{itemize}
    \vskip 4pt
\end{minipage} & 2 & 2 & 6 & 0 & 0 \\
  \hline
  4b & \begin{minipage}{3.0cm}
    \vskip 4pt
    Multi-GCS with half fixed times
    \vskip 4pt
\end{minipage} & \begin{minipage}{0.6in}
    \vskip 4pt
    \begin{itemize}
    	\item 2 MON
        \item 2 ES
        \item 2 TP
    \end{itemize}
    \vskip 4pt
\end{minipage} & \begin{minipage}{0.7in}
    \vskip 4pt
    \begin{itemize}
    	\item 1 HALE
        \item 1 MALE
        \item 1 UCAV
        \item 2 URAV
    \end{itemize}
    \vskip 4pt
\end{minipage} & 2 & 2 & 3 & 3 & 0 \\
  \hline
  4c & \begin{minipage}{3.2cm}
    \vskip 4pt
    Multi-GCS with half fixed times and dependencies
    \vskip 4pt
\end{minipage} & \begin{minipage}{0.6in}
    \vskip 4pt
    \begin{itemize}
    	\item 2 MON
        \item 2 ES
        \item 2 TP
    \end{itemize}
    \vskip 4pt
\end{minipage} & \begin{minipage}{0.7in}
    \vskip 4pt
    \begin{itemize}
    	\item 1 HALE
        \item 1 MALE
        \item 1 UCAV
        \item 2 URAV
    \end{itemize}
    \vskip 4pt
\end{minipage} & 2 & 2 & 3 & 3 & 1 \\
  \hline
  4d & \begin{minipage}{3.0cm}
    \vskip 4pt
    Multi-GCS with unfixed times
    \vskip 4pt
\end{minipage} & \begin{minipage}{0.6in}
    \vskip 4pt
    \begin{itemize}
    	\item 2 MON
        \item 2 ES
        \item 2 TP
    \end{itemize}
    \vskip 4pt
\end{minipage} & \begin{minipage}{0.7in}
    \vskip 4pt
    \begin{itemize}
    	\item 1 HALE
        \item 1 MALE
        \item 1 UCAV
        \item 2 URAV
    \end{itemize}
    \vskip 4pt
\end{minipage} & 2 & 2 & 0 & 6 & 0 \\
  \hline
  4e & \begin{minipage}{3.2cm}
    \vskip 4pt
    Multi-GCS with unfixed times and dependencies
    \vskip 4pt
\end{minipage}  & \begin{minipage}{0.6in}
    \vskip 4pt
    \begin{itemize}
    	\item 2 MON
        \item 2 ES
        \item 2 TP
    \end{itemize}
    \vskip 4pt
\end{minipage} & \begin{minipage}{0.7in}
    \vskip 4pt
    \begin{itemize}
    	\item 1 HALE
        \item 1 MALE
        \item 1 UCAV
        \item 2 URAV
    \end{itemize}
    \vskip 4pt
\end{minipage} & 2 & 2 & 0 & 6 & 3 \\
  \hline
  5 & \begin{minipage}{3.0cm}
    \vskip 4pt
    Complex mission with all assets at a time
    \vskip 4pt
\end{minipage} & \begin{minipage}{0.6in}
    \vskip 4pt
    \begin{itemize}
    	\item 2 MON
        \item 1 ES
        \item 2 TP
        \item 2 MAP
    \end{itemize}
    \vskip 4pt
\end{minipage} & \begin{minipage}{0.7in}
    \vskip 4pt
    \begin{itemize}
        \item 2 MALE
        \item 1 UCAV
        \item 2 URAV
    \end{itemize}
    \vskip 4pt
\end{minipage} & 3 & 3 & 4 & 3 & 1 \\
  \hline
\end{tabular}
}
\caption{Features of the different datasets designed.}
\label{tab:datasets}
\end{table}

\subsection{Experimental setup}\label{setup}
Table \ref{tab:parameters} shows the parameters used throughout the experimental phase, being $\mu + \lambda$ the selection criteria used, where $\lambda$ is the number of offspring (population size), and $\mu$ the elitism size, i.e. the number of the best parents that survive from current generation to the next. Each problem is run 10 times and the best of these 10 executions is selected.

\begin{table}[h]
\begin{center}
\begin{tabular}{|l||c|}
\hline
\textbf{Mutation probability} & 0.1\\
\hline
\textbf{Generations} & 300\\
\hline
\textbf{Population size} & 1000\\
\hline
\textbf{Selection criteria ($\mu + \lambda$)} & $100 + 1000$ \\
\hline
\textbf{Stopping criteria generations} & 10 \\
\hline
\end{tabular}
\caption{Experimental setup for the MOGA-CSP.}
\label{tab:parameters}
\end{center}
\end{table}

\subsection{Comparative Assessment of objective variables}
There are several parameters which can be used to measure the quality of a solution, such as the fuel consumption, the makespan, the cost of the mission, etc. As can be shown in section \ref{fitness}, this new algorithm considers 6 different optimization variables: number of UAVs, total flight time, total fuel consumption, total distance traversed, total cost and the makespan. To tune up the fitness function designed for the new algorithm, a comparative assessment of these variables is carried out. For this purpose the mission from dataset 1 has been chosen to study the behaviour of the algorithm according to the variables which are being optimized.

In these experiments, when obtaining several solutions in an execution, the average of the each optimization variable is computed. Then, in order to compare different executions (of different optimization variables), a weighted average over the values of the optimization variables is employed as rating value:

\begin{equation}
  Rating(sol) = \sum_{v \in OptVar}{\frac{v(sol)-min(v)}{max(v)-min(v)}}
\end{equation}

First, each variable is optimized individually. The results obtained for each optimization variable are shown in Table \ref{tab:oneOptimizationVar}. Analysing the results, it can be noticed that there are many different optimal solutions for the variables \textbf{Number of UAVs} and \textbf{Makespan}. In fact, none of them got to converge because new solutions were still being obtained at generation 300.

\begin{table*}[h]
\begin{center}
\scalebox{0.78}{
\begin{tabular}{|p{1.1cm}||c|c||c|c|c|c|c|c||c|}
\hline
\textit{Variable} & \textit{N. Sol.} & \textit{N. Gen.} & \textit{UAVs} & \textit{Fuel(L)} &\textit{F. Time(h)} & \textit{Dist.(NM)} & \textit{Cost} & \textit{Mak.(h)} & \textit{Rating} \\
\hline
Distance & 1 & \textbf{13} & 4 & 754.12 & 3.02 & \textbf{939.09} & \textbf{47.66} & 3.64 & \textbf{1.611} \\
\hline
F. Time & 4 & 14 & 4 & 771.08 & \textbf{3.02} & 949.74 & \textbf{47.66} & 3.64 & 2.031 \\
\hline
Cost & 4 & 14 & 4 & 771.08 & \textbf{3.02} & 949.74 & \textbf{47.66} & 3.64 & 2.031 \\
\hline
Fuel & 1 & \textbf{13} & 4 & \textbf{751.57} & 3.24 & 939.17 & 49.69 & 3.67 & 2.828 \\
\hline
Makespan & 1000 & \textit{300} & \textbf{3} & 810.06 & 3.21 & 1021.35 & 52.71 & \textbf{3.59} & 3.355 \\
\hline
UAVs & 1000 & \textit{300} & \textbf{3} & 798.44 & 3.52 & 1018.49 & 52.84 & 3.64 & 4.332 \\
\hline
\end{tabular}
}
\caption{Comparative assessment of optimization variables using each variable individually. The values of the optimization variables presented here are the average of their values in all the solutions obtained. The best result for each optimization variable is marked in bold.}
\label{tab:oneOptimizationVar}
\end{center}
\end{table*}

Regarding the rest of variables, it can be seen that optimizing the \textbf{cost} or \textbf{flight time} gave the same results. However, the \textbf{fuel consumption} and the \textbf{distance traversed} gave different results. In the case of the distance, it can be appreciated that it also got the best optimization value for Cost, and nearly optimal value for flight time (with a difference of $10^{-6}$ respect to the best value). In fact, optimizing the distance obtained the best rating, so it is a potential candidate to use in the fitness function of the MOGA-CSP approach.

Afterwards, the MOGA-CSP algorithm has been run optimizing each pair of the previous variables. The results obtained can be seen in Table \ref{tab:twoOptimizationVar}.

\begin{table}[h]
\begin{center}
\scalebox{0.78}{
\begin{tabular}{|p{1.1cm}||c|c||c|c|c|c|c|c||c|}
\hline
\textit{Variables} & \textit{N. Sol.} & \textit{N. Gen.} & \textit{UAVs} & \textit{Fuel(L)} &\textit{F. Time(h)} & \textit{Dist.(NM)} & \textit{Cost} & \textit{Mak.(h)} & \textit{Rating} \\
\hline
Distance & \multirow{2}{*}{3} & \multirow{2}{*}{16} & \multirow{2}{*}{4} & \multirow{2}{*}{757.57} & \multirow{2}{*}{3.02} & \multirow{2}{*}{939.32} & \multirow{2}{*}{47.66} & \multirow{2}{*}{3.64} & \multirow{2}{*}{\textbf{1.396}} \\
F. Time & & & & & & & & &  \\
\hline
Distance & \multirow{2}{*}{4} & \multirow{2}{*}{15} & \multirow{2}{*}{4} & \multirow{2}{*}{759.39} & \multirow{2}{*}{3.02} & \multirow{2}{*}{941.50} & \multirow{2}{*}{47.66} & \multirow{2}{*}{3.64} & \multirow{2}{*}{1.450} \\
Cost & & & & & & & & &  \\
\hline
Distance & \multirow{2}{*}{2} & \multirow{2}{*}{15} & \multirow{2}{*}{3.5} & \multirow{2}{*}{769.57} & \multirow{2}{*}{3.10} & \multirow{2}{*}{965.57} & \multirow{2}{*}{49.48} & \multirow{2}{*}{3.61} & \multirow{2}{*}{1.681} \\
Makespan & & & & & & & & &  \\
\hline
Distance & \multirow{2}{*}{2} & \multirow{2}{*}{17} & \multirow{2}{*}{3.5} & \multirow{2}{*}{769.57} & \multirow{2}{*}{3.10} & \multirow{2}{*}{965.57} & \multirow{2}{*}{49.48} & \multirow{2}{*}{3.61} & \multirow{2}{*}{1.681} \\
UAVs & & & & & & & & &  \\
\hline
F. Time & \multirow{2}{*}{4} & \multirow{2}{*}{15} & \multirow{2}{*}{4} & \multirow{2}{*}{771.08} & \multirow{2}{*}{\textbf{3.02}} & \multirow{2}{*}{949.74} & \multirow{2}{*}{\textbf{47.66}} & \multirow{2}{*}{3.64} & \multirow{2}{*}{1.729} \\
Cost & & & & & & & & &  \\
\hline
Cost & \multirow{2}{*}{13} & \multirow{2}{*}{14} & \multirow{2}{*}{4} & \multirow{2}{*}{753.50} & \multirow{2}{*}{3.11} & \multirow{2}{*}{939.17} & \multirow{2}{*}{48.47} & \multirow{2}{*}{3.66} & \multirow{2}{*}{1.741} \\
Fuel & & & & & & & & &  \\
\hline
Distance & \multirow{2}{*}{20} & \multirow{2}{*}{14} & \multirow{2}{*}{4} & \multirow{2}{*}{\textbf{752.67}} & \multirow{2}{*}{3.12} & \multirow{2}{*}{\textbf{939.12}} & \multirow{2}{*}{48.50} & \multirow{2}{*}{3.66} & \multirow{2}{*}{1.745} \\
Fuel & & & & & & & & &  \\
\hline
F. Time & \multirow{2}{*}{12} & \multirow{2}{*}{15} & \multirow{2}{*}{4} & \multirow{2}{*}{753.48} & \multirow{2}{*}{3.11} & \multirow{2}{*}{939.18} & \multirow{2}{*}{48.58} & \multirow{2}{*}{3.66} & \multirow{2}{*}{1.784} \\
Fuel & & & & & & & & &  \\
\hline
Fuel & \multirow{2}{*}{4} & \multirow{2}{*}{15} & \multirow{2}{*}{3.5} & \multirow{2}{*}{767.52} & \multirow{2}{*}{3.25} & \multirow{2}{*}{965.62} & \multirow{2}{*}{50.76} & \multirow{2}{*}{3.63} & \multirow{2}{*}{2.179} \\
Makespan & & & & & & & & &  \\
\hline
Cost & \multirow{2}{*}{9} & \multirow{2}{*}{18} & \multirow{2}{*}{3.11} & \multirow{2}{*}{805.22} & \multirow{2}{*}{3.16} & \multirow{2}{*}{1004.30} & \multirow{2}{*}{50.89} & \multirow{2}{*}{3.59} & \multirow{2}{*}{2.529} \\
Makespan & & & & & & & & &  \\
\hline
Cost & \multirow{2}{*}{9} & \multirow{2}{*}{16} & \multirow{2}{*}{3.11} & \multirow{2}{*}{805.22} & \multirow{2}{*}{3.16} & \multirow{2}{*}{1004.30} & \multirow{2}{*}{50.89} & \multirow{2}{*}{3.59} & \multirow{2}{*}{2.529} \\
UAVs & & & & & & & & &  \\
\hline
Makespan & \multirow{2}{*}{\textit{1000}} & \multirow{2}{*}{\textit{300}} & \multirow{2}{*}{\textbf{3}} & \multirow{2}{*}{808.92} & \multirow{2}{*}{3.21} & \multirow{2}{*}{1020.69} & \multirow{2}{*}{52.65} & \multirow{2}{*}{\textbf{3.58}} & \multirow{2}{*}{3.046} \\
UAVs & & & & & & & & &  \\
\hline
F. Time & \multirow{2}{*}{18} & \multirow{2}{*}{16} & \multirow{2}{*}{3.11} & \multirow{2}{*}{815.64} & \multirow{2}{*}{3.12} & \multirow{2}{*}{1027.05} & \multirow{2}{*}{51.97} & \multirow{2}{*}{3.59} & \multirow{2}{*}{3.121} \\
Makespan & & & & & & & & &  \\
\hline
F. Time & \multirow{2}{*}{18} & \multirow{2}{*}{20} & \multirow{2}{*}{3.11} & \multirow{2}{*}{815.64} & \multirow{2}{*}{3.12} & \multirow{2}{*}{1027.05} & \multirow{2}{*}{51.97} & \multirow{2}{*}{3.59} & \multirow{2}{*}{3.121} \\
UAVs & & & & & & & & &  \\
\hline
Fuel & \multirow{2}{*}{2} & \multirow{2}{*}{14} & \multirow{2}{*}{3.5} & \multirow{2}{*}{757.53} & \multirow{2}{*}{3.87} & \multirow{2}{*}{977.82} & \multirow{2}{*}{51.96} & \multirow{2}{*}{3.75} & \multirow{2}{*}{3.879} \\
UAVs & & & & & & & & &  \\
\hline
\end{tabular}
}
\caption{Comparative assessment of optimization variables using each pair of variables. The values of the optimization variables presented here are the average of their values in all the solutions obtained. The best result for each optimization variable is marked in bold.}
\label{tab:twoOptimizationVar}
\end{center}
\end{table}

In these results, it is appreciable that the two best combinations obtained according to the rating (i.e. the optimization of the distance and the flight time, and the distance and the cost) obtained good results for four of the variables (the cost, the distance, the fuel and the flight time) but poor results for the rest (the makespan and the number of UAVs). On the other hand, the third and fourth best combinations according to the rating (i.e. the optimization of the distance and the makespan, and the distance and the number of UAVs) obtained medium results for all the variables.

So, in order to find good solutions optimizing all the variables, the variables selected to optimize are the \textbf{distance} and the \textbf{makespan}, which gave medium values for all the optimization variables. Other possibility would have been using the number of UAVs instead of the makespan, but as the makespan is a float value, it will be better for optimizing problems with very similar solutions (e.g. all the best solutions when optimizing any variable uses 1 UAV because the mission can be performed with just one and other available UAVs are far away from the tasks of the mission).

Finally, the MOGA-CSP algorithm is executed with this problem trying to optimize all the six objective variables, and the results obtained can be seen in Table \ref{tab:allOptimizationVar}. As can be seen, the average obtained here for all objective variables are worst than the ones obtained in the previously proposed combination of distance and makespan, as well as the rating value. This corroborates the assumption of selecting this combination, which will be used in the fitness of the MOGA-CSP in the next experiment when solving the different datasets.

\begin{table*}[h]
\centering
\scalebox{0.78}{
\begin{tabular}{|p{1,05cm}||c|c||c|c|c|c|c|c||c|}
\hline
\textit{Variable} & \textit{N. Sol.} & \textit{N. Gen.} & \textit{UAVs} & \textit{Fuel(L)} &\textit{ F. Time(h)} & \textit{Dist.(NM)} & \textit{Cost} & \textit{Mak.(h)} & \textit{Rating} \\
\hline
All & 46 & 20 & 3.52 & 772.76 & 3.15 & 970.13 & 50.13 & 3.62 & 2.077  \\
\hline
\end{tabular}
}
\caption{Comparative assessment of optimization variables using all variables. The values of the optimization variables presented here are the average of their values in all the solutions obtained.}
\label{tab:allOptimizationVar}
\end{table*}

\subsection{Evaluation of the algorithm results}
Once the fitness function of the algorithm has been tuned up, and the better optimization variables (distance and makespan) have been selected, the MOGA-CSP algorithm is tested using them for each dataset described in Table \ref{tab:datasets}. To evaluate the results obtained, the real POF of each dataset is computed using MOBB algorithm. Then, it is compared with the solutions provided by the new MOGA-CSP approach using the hypervolume metric. As was mentioned in section \ref{gas}, when the hypervolume is 0, the obtained solutions are optimal. On the other hand, as this value increases, the result obtained distances from the optimal result. Table \ref{tab:results} shows the results obtained from this experiment. This table presents the hypervolumes obtained and the number of generations needed to converge for each dataset.

\begin{table}[h]
\centering
\begin{tabular}{|c||c|c| }
  \hline
  Problem & Hypervolume & Generations \\
  \noalign{\hrule height 2pt}	
  1 & 0 & 15 \\
  \hline
  2 & 0 & 14 \\
  \hline
  3 & 0 & 12 \\
  \hline
  4a & 0 & 39 \\
  \hline
  4b & 0 & 43 \\
  \hline
  4c & 0 & 56 \\
  \hline
  4d & 0.99 & 57 \\
  \hline
  4e & 0 & 62 \\
  \hline
  5 & 0.01 & 122 \\
  \hline
\end{tabular}
\caption{Hypervolume and number of generations needed for convergence of the MOGA-CSP solver for the 9 MPP datasets provided.}
\label{tab:results}
\end{table}

\begin{figure}[hbt]
\centering
\includegraphics[width=\linewidth]{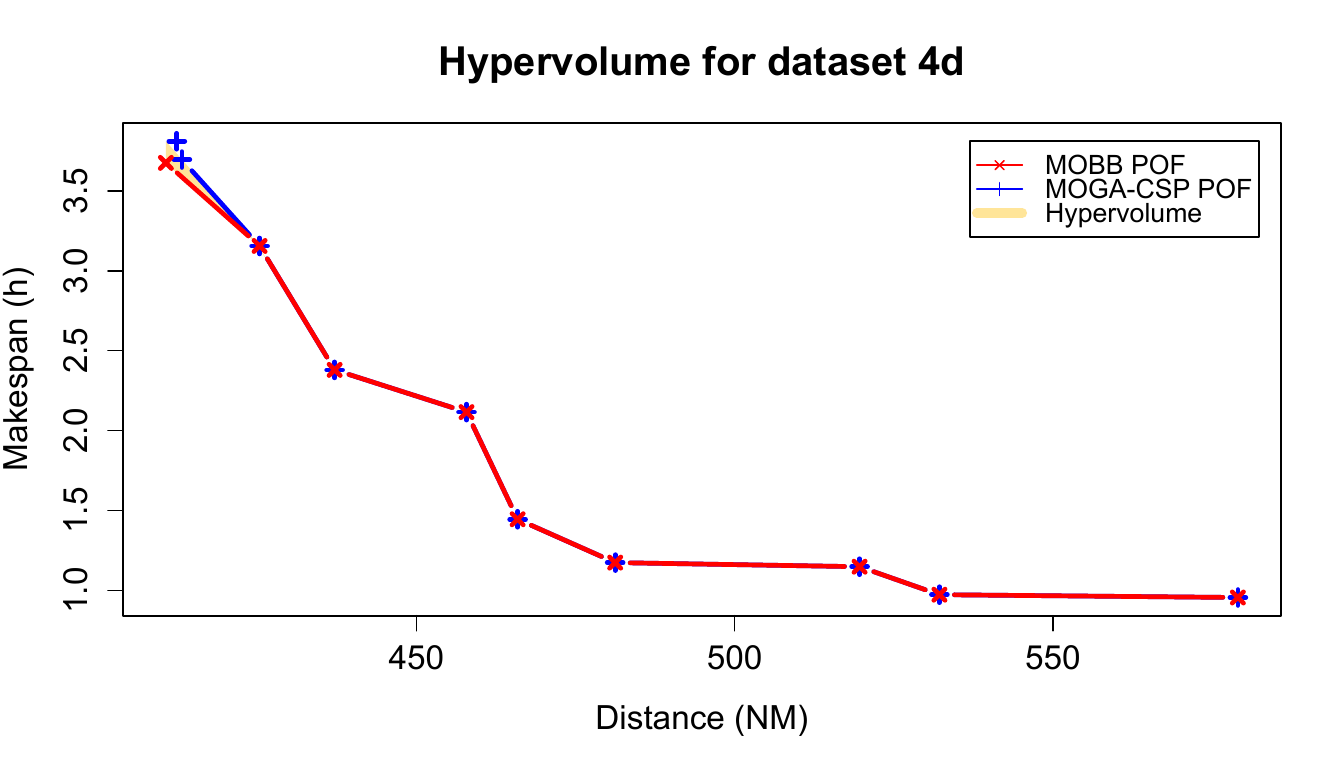}
\caption{Hypervolume for the solutions obtained with MOGA-CSP optimizing the distance and the makespan in dataset 4d.}
\label{fig:hypervolume}
\end{figure}

As can be appreciated, the datasets with 1 GCS (from 1 to 3) converge very fast, independently of the NFZs needed to avoid or the Multi-UAV tasks. On the other hand, the datasets with 2 GCS (from 4 to 5) converge near generation 50, being easier to converge the problems with more fixed times tasks and harder for the problems with more unfixed tasks. The dataset 4d did not get a hypervolume so good as the others datasets. Figure \ref{fig:hypervolume} represents the Distance vs. Makespan POFs for the solutions obtained with both algorithms (MOGA-CSP and MOBB) in this problem. There, it is appreciable that the MOGA-CSP approach did not get to obtain the best solution optimizing the distance (left of the POF), and this made the hypervolume (represented in yellow) higher in this problem.

Finally, the complex solution with 7 tasks, 5 UAVs and 3 GCSs, i.e. mission 5, got to converge at generation 122, and its hypervolume resulted quite good, very close to the optimum POF.

In conclusion, the MOGA-CSP approach with the distance and the makespan used as optimization variables approximates quite well the POF in most cases. As the problem becomes more complex, specially in number of GCSs, the algorithm needs more generations to converge.

\section{Discussion}\label{conclusions}
In this paper, the Multi-UAV Mission Planning Problem has been presented, with a special consideration as a Multi-GCS problem. This problem involves a complex characteristic management during the process of task assignment, such as task dependencies, NFZ avoidance and time computations.

The problem has been modelled as a Temporal Constraint Satisfaction Problem, where six sets of variables have been considered for the CSP: the task assignments, the orders, the GCS assignments, the path flight profiles, the return flight profiles and the sensors used. On the other hand, a wide range of constraints have been presented, including GCS constraints, temporal constraints, dependency constraints, autonomy constraints or distance constraints among others.

Finally, a hybrid MOGA-CSP approach has been presented for solving the MPP problem. This approach uses a fitness function divided in two phases. Firstly, a penalty function uses the CSP to check if the solutions are valid. Then a multi-objective function tries to approach the Pareto Optimal Frontier of the problem minimizing the optimization variables (number of UAVs employed, makespan of the mission, total fuel consumption, etc...). In addition, the crossover and mutation operators, and also the stopping criteria have been specifically designed and implemented for this approach.


The experiments presented have been performed using varied datasets of different complexity. First, a comparative assessment of the optimization variables has been performed in order to tune up the fitness function designed. The results show that the best combination in order to obtain good results for all the variables is optimizing the distance and the makespan.


Afterwards, the MOGA-CSP approach using the previous combination of variables in the fitness function has been tested with all the datasets designed. Analysing the experimental results, it can be seen that the MOGA-CSP algorithm obtains good results for all the proposed datasets, converging to the optimal POF in most of them. Nevertheless, as the problems become more complex, the MOGA-CSP approach needs more generations to reach an optimal or near-optimal solution. In order to outperform these results, it can be interesting to extend the new approach applying some constraints in the operators of the GA in order to avoid some invalid solutions before the CSP check.

In future works, the approach will be compared using other Multiobjective Algorithms, such as SPEA2, in order to find the best performing combination for this approach. In order to find an optimum configuration, a Meta-Evolutionary algorithm will be implemented and used to optimize the different parameters of the approach. On the other hand, this problem will be extended adding a decision making layer that will interact with a UAV Mission operator. This new feature will allow the operator to decide which variables must be optimized and which of the obtained solutions are the most suitable.

\begin{acknowledgements}
This work is supported by: Spanish Ministry of Science and Education and Competitivity and European Regional Development Fund (FEDER) under project TIN2014-56494-C4-4-P, Comunidad Aut\'onoma de Madrid under project CIBERDINE S2013/ICE-3095 and Savier Project (Airbus Defence \& Space, FUAM-076915). The authors would like to acknowledge the support obtained from Airbus Defence \& Space, specially from Savier Open Innovation project members: Jos\'e Insenser, C\'esar Castro, Gemma Blasco and In\'es Moreno.

\end{acknowledgements}

\bibliographystyle{spmpsci}      
\bibliography{biblio}   

\end{document}